# ProjB: An Improved Bilinear Biased ProjE Model for Knowledge Graph Completion

Mojtaba Moattari [*], Sahar Vahdati[+], Farhana Zulkernine[*]

mojtaba.moattari@queensu.ca, vahdati@infai.org, farhana.zulkernine@queensu.ca

[*] School of Computing, Queen's University, Kingston, Ontario, Canada

[+] The Institute for Applied Informatics, Leipzig University, Leipzig, Germany

## ABSTRACT

Knowledge Graph Embedding (KGE) methods have gained enormous attention from a wide range of AI communities including Natural Language Processing (NLP) for text generation, classification and context induction. Embedding a huge number of inter-relationships in terms of a small number of dimensions, require proper modeling in both cognitive and computational aspects. Recently, numerous objective functions regarding cognitive and computational aspects of natural languages are developed. Among which are the state-of-the-art methods of linearity, bilinearity, manifold-preserving kernels, projection-subspace, and analogical inference. However, the major challenge of such models lies in their loss functions that associate the dimension of relation embeddings to corresponding entity dimension. This leads to inaccurate prediction of corresponding relations among entities when counterparts are estimated wrongly. ProjE model is improved in this work regarding all translative and bilinear interactions while capturing entity nonlinearity . Experimental results on benchmark Knowledge Graphs (KGs) such as FB15K and WN18 show that the proposed approach outperforms the state-of-the-art models in entity prediction task using linear and bilinear methods and other recent powerful ones. In addition, a parallel processing structure is proposed for the model in order to improve the scalability on large KGs. The effects of different adaptive clustering and newly proposed sampling approaches are also explained which prove to be effective in improving the accuracy of knowledge graph completion.

## 1 Introduction

With the advent of Knowledge Graph Embedding models (KGEs), several AI-based tasks such as link prediction, question answering, Natural Language Processing (NLP) have had breakthrough in results [41]. In NLP, the recent embedding-based advances in extraction of interrelations among terms have led to more geometrically justifiable concept mapping solutions and relation extraction approaches [2]. Generally, the strength of Knowledge Graph (KG) technology belongs to the semantics expressed in the underlying ontologies, axioms and the multi-relational representation of information as entities (nodes) and relations (links) known as triples in the form of <h, r, t> depicting (head entity, relation, tail entity) . The KGE models provide low-dimensional representation of the triples as **<h, r, t>**, where **h** accounts for a head entity vector, **r** associates a relationship vector, and **t** represents a tail entity vector.

During the past years, explosion of deep learning methods together with the rise of embedding-based models have impacted the development of more generalizable language technologies [47]. Such models are capable of handling unseen data, transferring information from one domain to another, and cooperating with rule-based reasoning and classical inference methods. Models based on graph embedding are not only used to improve the performance of search engines and recommender systems, but also are used as methods for creation of priori knowledge in Question Answering and Machine Translation (MT) [3] services. There is a growing interest among NLP researchers to develop Multi-Task Learning (MTL) frameworks that restrict corpus embedding context to domain of ontologies. Such ontologies may be related to Knowledge Graphs (KG), sets of parse trees, corpuses from different languages or even visual features of multi-modal information [4-9].

Despite the great impact, one of the most crucial challenges in KG embedding is finding meaningful representations which are capable of accurately estimating relations between entities in presence of unseen entities, i.e. Knowledge Graph Completion (KGC) [11]. Such capability of generalizing to unseen, missing, or out of vocabulary terms, not only necessitates a promising generalization power from the model, but also generates semantically relevant representations useable for pre-training in applications of interest [12]. Consequently, KGC (entity prediction) is the task that we focus in the evaluation of our work. Entity prediction task aims at extracting previous/new entities given an entity and its relation.

The most basic forms of KGs have begun from the TransE model proposed by Bordes et al. [13] where the head entity is added to a corresponding relation with a purpose that the ultimate result leads to the tail entity. The follow up models attempted to cover the problem of TransE in encoding of relational patterns and structure preservation. With recent advances, the current KGE models are capable of handling complex structures such as hierarchical and structural aspects of human nervous system. Among more recent models, TtransE and PtransE can be mentioned, that respectively learn entities interrelations in pathways using Recurrent Neural Nets (RNN), and triple of entities structure [13, 14]. Afterwards, Neural Tensor Network (NTN) and Computational Vector Space Models (CVSM) have been proposed to transfer simplistic unstructured KGs to NN-based representation [15, 16]. However, all of these approaches regard one relationship at a time and mostly fail on larger real-world settings. In models oriented towards Computational Geometry such as ManifoldE and ANALOGY [44, 45], respectively, learn Hilbert spaces to control nonlinearity and smoothness with which entities interrelate, and regularize to enforce analogical interaction between graph entities. Despite the salient outperformance, these models have excessive parameters, highly complex structure, and outraging complexity growth over large scale graphs. Although other models like DKRL improved the state-of-the-art performance using Convolutional Neural Networks (CNN), they need pre-computation of nodes [17]. Therefore, they are not self-contained.

In this paper, we present a comprehensive revision of the ProjE both to improve the embedding performance and the scalability of parallel computation. ProjE, a KGE developed by Shi and Weninger [12], learns heads and relations such that for each tail, head feature vector is reweighted by the sum of the sigmoid output of its corresponding relation/entity. The sigmoid function of this model works similar to an attention mechanism, bolding those dimensions of input features which have more contribution to the output. The transformation therefore is nonlinear, the selection of promising input dimensions is flexible and its complexity is close to TransE which were the reasons for choosing ProjE to build on top. Along with the aforementioned advantages of ProjE, the model has some deficiencies as described below:

- It corrupts the whole entities/relations in one embedding dimension in the case of getting a relation/entity caught into misleading local optima. To prove that, a hypothesis testing is performed on ProjE and the proposed method in Section 3.6.
- The model learns to reweight head dimensions by relations, but it neither reweights tails dimensions using heads nor reweights by relations.
- The model does not prune unnecessary dimensions in relation embedding constrained on the head in context, therefore leading to decrease in the performance of tails-scoring.
- It is flexible to be extended using tensor based parallel or distributed processing.
- Finally, ProjE neither handles bilinear interactions of entity and relation vectors (i.e., entity-entity, entity-relation interactions), nor gives heads the chance to learn independently without affecting the relations they interact with (i.e., entity-learnable bias, relation-learnable bias interaction).

Therefore, in this paper we propose ProjB, a bilinear ProjE, which regards all interactions between relations and entities. It works as an attention mechanism to reweight input elements' contributions to each tail in the output. It acts more robustly than ProjE by reducing wrongly trained entity/relation effects on correctly-trained relation/entity. As a result, it improves point-wise and list-wise raw scores in FB15K/WN18 test data's entity prediction. Furthermore, it controls the extent of linearity and bilinearity by learning two reweighting terms. The drawbacks of the aforementioned models are addressed and elaborated in the upcoming sections. Hence, the contribution of this paper are as follows:

- A novel bilinear multi-biased KGE model dubbed ProjB is proposed which is developed on top of the ProjE model to improve the deficiencies of several state-of-the-art methods in entity and link prediction.
- A new feature engineering procedure is discussed and the comparison results demonstrate newly proposed sampling and feature selection techniques as useful tools for the researchers.
- A new variation of the proposed model has been constructed based on Tensor Calculus to improve the running speed and scalability aspects of the algorithm.

## 2 Related Work

In this section, we provide an overview of the basic linear and bilinear KGEs to further unravel their limitations. We have refrained from discussing more complex linear and bilinear KGE models which are out of scope of this work. Among the most basic and general form of KGE models is Neural Tensor Network (NTN), proposed in 2013 by Socher et al. [15], which enforces to learn projections of all relations along with both head and tail entities. This is performed with extra bilinear interaction between corresponding subspaces of head and tail entities. The score function of NTN is given in the Eq. 1

$$E(h, r, t) = \boldsymbol{u}_r^T f(\boldsymbol{h}^T \boldsymbol{W}_r \boldsymbol{t} + \boldsymbol{W}_{rh} \boldsymbol{h} + \boldsymbol{W}_{rt} \boldsymbol{t} + \boldsymbol{b}_r) \qquad (1)$$

where $\boldsymbol{u}_r$, $\boldsymbol{W}_r$, $\boldsymbol{W}_{rh}$, and $\boldsymbol{W}_{rt}$ are all relationship-specific parameters. $f$ is Sigmoid activation function. $\boldsymbol{W}_r \in \mathcal{R}^{z*z*k}$ is a tensor, which when multiplied in the bilinear tensor product $\boldsymbol{h}^T \boldsymbol{W}_r^{1:k} \boldsymbol{t}$, leads to a k-dimensional vector, the same dimensions as the relation vector $\boldsymbol{r}$. $\boldsymbol{W}_{rh}$, and $\boldsymbol{W}_{rt}$ are two dimensional matrices with dimensions $k$ and $z$ respectively. The embedding vectors $\boldsymbol{h}$, $\boldsymbol{t}$, and $\boldsymbol{b}_r$ correspond to heads, tails and relationships between them respectively, each of which have $z$, $z$, and $k$ as their number of dimensions. The bilinear term demands $n_r z k 2$ number of parameters where terms containing $W_{rx}$ have a parameter count of $n_x z k$ (here $n_x$ is the total number of *x*, and *x* stands for either *h* or *t* i.e., heads or tails respectively. Therefore, $n_h$ and $n_t$ are respectively the total number of heads and tails.). This model is by far the most computationally complex score function among the linear and bilinear KGE models, imposing a lot of burden in large-scale graph processing and making embeddings impractical for real use case scenarios. To tackle the issue of computational burden in NTN, Bordes et al. proposed TransE, by constraining NTN's W matrices as independent on specific relations [13]. The score function of TransE is known for its simplicity, as shown in Eq. 2.

$$E(h,r,t) = ||h + r - t|| \qquad (2)$$

Where **h, r, t** are head, relation and tail embedding vectors respectively. TransE was reported not to be capable of encoding the patterns of relations. For improving this problem of TransE, a chain of models has been proposed one after the other. Among them, we can name TransH that considered orthogonality constraint [18], TransR that changed and learned hyperplanes for relations to be translated to projected entities [19], and HolE which used other operators instead of subtracting heads from tails in a translated way [20]. However, all of them are linear, disregarding bilinear interactions between relations and entities. Bilinear interactions contain information that can more easily supervise relational instances of entity vectors with respect to each other.

To suggest a cure for NTN complexity and the lack of bilinear relationships in TransE, a holistic framework called DistMult [21] was proposed by Yang et al. DistMult functions by relaxing the dependency of bilinear matrix on relation index. The general framework for scoring the entity interactions is given in Eq. 3 and Eq. 4.

$$E(h,r,t) = -\left(2g_r^a(y_{e_1}, y_{e_2}) - 2g_r^b(y_{e_1}, y_{e_2}) + s||V_r||_2^2\right) \quad (3)$$

$$g_r^a(y_{e_1}, y_{e_2}) = A_r^T \begin{pmatrix} y_{e_1} \\ y_{e_2} \end{pmatrix} \qquad g_r^b(y_{e_1}, y_{e_2}) = y_{e_1}^T B_r y_{e_2} \quad (4)$$

where $A_r \in R^{k \times 2z}$, $y_{e1}$; $y_{e2}$ are two entity vectors (heads and tails embedding vectors described in Eq. 1), z is the number of dimensions for heads and tails embedding vectors. By associating different matrices to $A_r^T$ and $B_r$, Yang et al. proved that variety of multi-relational models (Distance-based [21], Single-Layer [21], TransE [13], NTN [15]) can be derived. s in Eq. 3 is 0 for all models except for TransE which is 1 in that case. The term $||V_r||_2^2$ is appears by expanding L2 norm in Eq. 2.

As it is evident from formulation, $g_r^a$ is a linear map of $y_{e_1}, y_{e_2}$ and the function $g_r^b$ is a bilinear map between $y_{e_1}, y_{e_2}$. $M_r$ does not depend on any specific relation. Although reduction of parameters out of this approach improves link prediction over large-scale graphs such as FB15K [10], yet it disregards interaction between each entity dimension with all dimensions of a relation embedding-vector. This bilinearity type only favors correlations of linear dependent embedding dimensions of entities together, without regarding the inter-relations among all dimensions. This makes respective bilinearity constraint a weak kind. In the proposed approach, we have provided a strong form of bilinearity out of additional dimensional interactions.

In Eq. 3, the function g is for network prediction and assigns scores to each of the tails lied in the desired head and relation of the graph tensor. Graph tensor is created out of the entity and relation indices in the graph. When graph tensors are binary, the network's ground-truth outputs are 1 or 0; while for real-valued outputs, the outputs are non-negative real numbers.

Another approach that focuses on KG completion is proposed by Shi and Weninger named ProjE that tackles the problems concerning the high computational complexity of NTN. This is done while removing the need for pre-training. The method avoids adding extra large number of transformation matrices by using a shared representation. The proposed model architecture is:

$$h_i(e,r) = g[W_{i,:} \, t + b_p], \; t = f(D_e e + D_r r + b_c), \qquad (5)$$

where, $f$ and $g$ are Sigmoid activation functions, which are identical across the paper, $h_i$ is the score for the predicted tail. $W_{i,:} \in R^{s \times k}$ is the candidate-entity matrix being derived by selecting the embedding vectors of entities. $b_p$ is the projection bias, and **s** is the number of candidate-entities. k is the number of dimensions of entity and relation embedding vectors. Variable i accounts for the index of tail. The matrices $D_r$ and $D_e$, represent engineered features corresponding to relations and heads respectively. $D_e$ and $D_r$ are $k \times k$ diagonal matrices, which serve as global entity and relationship weights respectively, and $b_c \in R^k$ is the combination bias. Contrary to previous formulas, we refrained to use the notation t, as the input to this model is both head and tail entities. As our proposed Knowledge Graph Embedding model is mainly designed for entity-linking, heads and tails can both be fed as e.

ProjE has a number of limitations such as being error-prone, not regarding different embedding parametrization for different cluster localities of entities/relations, not having a bilinear behavior, and disregarding relation-based attention. We propose a model to address these problems. To deal with NTN complexity, translational model linearity, weak bilinearity of DistMult (bilinear-diag), and avoid fine-grained interactions in ProjE, a bilinear variation of ProjE with cluster-related embedding vectors is proposed in the next section.

## 3  ProjB: Bilinear ProjE

We introduce bilinear biased version of **ProjE**, named as **ProjB**. The proposed framework governs the same methodology as ProjE for the design of score function. The methodology includes using scores for correlating a vector of all possible tails per each pair of (head, relation) as duals. This is in contrast to the previous approaches such as NTN or TransE where computations are performed in triples (head, relation, tail). This way of tackling duals instead of triples reduces a huge number of complexities and parameters, and thus improves parallel processing speed of model training. It is noteworthy to include that the top-ranked candidates are regarded as the correct tails in the task of knowledge graph completion while the low-ranked ones are considered as false tails. In this way, entity prediction is achieved by the model that facilitates the completion even in the presence of unseen entities.

## 3.1 ProjB Model Architecture

The algorithm is designed such that it handles the previous bilinear and linear drawbacks aforementioned in Section 2. One major challenge is to reduce the effect of possible embedding vectors that are wrongly trained. For this, a possible solution is to consider a larger number of dimension interactions. More precisely, the assumption is that one of the corresponding dimensions for a head vector has been caught in local optima during Gradient update. In the review of other KGE models that were mentioned previously, occurrence of such a vector possibly misleads a set of nearby relations and entities. By allowing all dimensions of heads to be affected by all dimensions of relations during training phase, it becomes more probable that the optimal dimensions eventually correct the suboptimal dimensions. Eq. 6 depicts the theoretical basis of this solution. Another challenge is to control the extent of linearity and bilinearity, so that the model decides to behave in a bilinear way only if the input graph demands. This problem is addressed by learning the context independent graph parameters that reweight the linear and bilinear terms. To achieve that through a score function without injecting huge complexity, the following formulation is derived. The objective is to embed the space of $e$ and $r$ in:

$$h_i(e, r) = g(W_{i,:} t + b_p) \quad (6)$$

$$t = f\left(\left(D_e e + b_{p_{c_e}}\right)\left(D_r r + b_{q_{c_r}}\right)^T\right) r \quad (7)$$

where f and g are activation functions. $h_i$ is the predicted score for the tail in the context of all entities[1], which is fed to the loss function in (8) for cross entropy minimization. e, t, and r are respectively the embedding results for head, tail entities and the relationships between them. $b_p$ denotes a global trainable constant, $p_{c_e}$ is the learned vector for clusters of entity groups, and $q_{c_r}$ is the learned vector for clusters of relation groups. Cluster-related embedding indices as p and q, are provided mainly to reduce parameters and preserve fine grained graph information. Contrary to the ProjE formulation, here $W_{i,:}$ shows the entity embedding vector (one column of WE, with dimensions s×k as mentioned for Eq. 5). $D_e$ corresponds to the entity vector input, and $D_r$ denotes the relation vector input. $D_e$, $D_r$ are derived by diagonalizing the engineered features in a square matrix. Engineered features are either co-occurring vectors or cluster densities which are described in the next section. In ProjE model, W represents the list of entity vectors, demanding a more complex differentiation in Gradient Descent. Each row of W corresponds to a specific entity, derived by counting the number of tails-relations co-occurring for the involved entity. The row vectors of W are initialized randomly by uniform distribution. Finally, the biases $b_{p_{c_e}}$ and $b_{q_{c_r}}$ are added to each sub-term to control weights of $D_e e$ and $D_r r$ in the ProjE model as shown in Eq. 7. The biases are different for each entity cluster $c_e$ and relation cluster cr. Both $c_e$ and $c_r$ are the cluster index of entity and relations respectively.

Multiplication of embedding vector $W_{i,:}$ by parameter vector t followed by a shift $b_p$ leads to hi, predicted score of a tail with all existing entities. The distance measure between hi and $\hat{h}_i$ is a cross-entropy function that optimizes the embedding vectors $W^E$ and $W^R$.

We shall note that when a new graph item appears (e.g. entity and its relations), the co-occurrence vectors of these entities and relations are fed into the model. Furthermore, both $c_e$ and $c_r$ are extracted by measuring the distance of entity and relation embedding for each cluster embedding. In the next step, such cluster embeddings are updated by averaging over each sub-item embedding per every update during training iterations.

For the testing process, the output hi shows prediction of tail scores on every entity of the graph. We select the entity with highest score as tail. The final loss function regarding the clusters cost is provided in Eq. 17 which will be discussed in the Subsection 3.1.1. By expanding the parenthesis in f input function of the Eq. 7, we not only model the entity-relation interrelations by cluster-dependent learnable weights, but also reveals the possible effect of entities and relations on the output score values without disturbance of one another. The reader can refer to Section 3.7 for more details. Fig.1 shows the structure of proposed model with some information regarding each parameter type. It starts with importing the inputs. The input is fed into the process using engineered features from top-left and top-right side of the figure, and $h_i$(e, r) is manifested in the bottommost side of it. Compared to Fig. 1 in [12], in this architecture it is evident that irrespective of the learnable biases added to both entity and relation embedding, an attention framework is added to relation-entity dimensional interaction matrix. This approach improves robustness of the algorithm as well as the accuracy of knowledge-graph completion.

---
[1] Score of regarding tail as each of entities

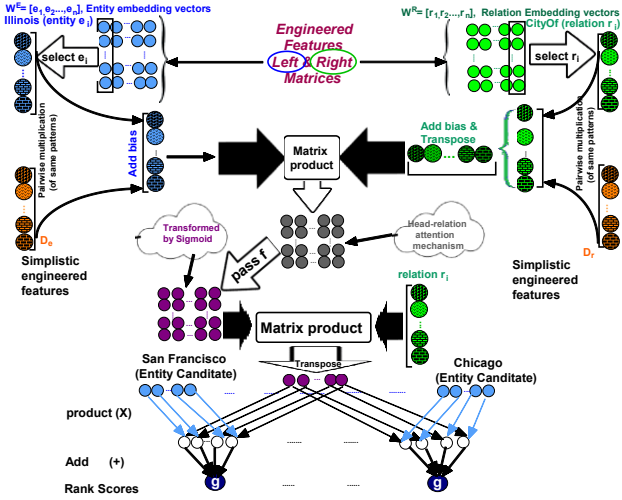

*Figure 1: Architecture and schematics of the proposed ProjB Method*

### 3.2 Finalized Loss Function

Enforcing the entity and relation embedding to move towards clusters centers, improves the fine-grained complexity of the corresponding vectors. It also causes them to assimilate to best sought similar contexts. Each cluster generally acts like a term/word sense, determining roles in directing the embedding vectors to more suitable states. The regularization to assimilate entities in the same cluster closer to the center is implemented as shown in Eq. 8.

$$R(e, r) = \sum_i V(\{e^i | e \in c_e\}) + V(\{r^i | r \in c_r\}) \quad (8)$$

Here, V is variance function, $e_i$ is the entity value at embedding dimension i, $\{e^i | e \in c_e\}$ is the set of all entity embedding vectors that belong to the cluster index $c_e$. The same routine holds for relation embeddings where we consider them as $\{r^i | r \in c_r\}$. The final objective function to minimize is shown in Eq. 9.

$$L = CE(h_i(e, r), \hat{h}_i(e, r)) + \delta R(e, r) \quad (9)$$

In this equation, CE is a cross-entropy function, δ is fine-tuned from the set {0.1, 0.01, 0.001, 0.0001} which is set to 0.001, as it provides the highest validation accuracy. Note that the clusters optimized here are only for assigning different embedding biases to different entity/relation clusters $b_{p_{c_e}}$ and $b_{q_{c_r}}$. The cluster assignment to entities and relations is done from the result of feature engineering, as described in the next subsection.

### 3.3 Feature Engineering Approach

The conventional trend of feature engineering approaches is to use different feature vectors for relations and entities, where for each relation or entity, the frequency of triples in the fixated relation/entity is counted, leading to a large vector of entity-relation co-occurrence matrix. The engineered features are provided in triples according to the graph connections, and then are fed to the model for training. This approach creates large vectors for entities/relations, necessitating dimension reduction methods like Principal Component Analysis (PCA) to reduce dimensions. Such blind approaches remove a lot of instructive information from data, causing several negative effects on results [48]. This conventional approach is used for comparison as the baseline approach in Table 4.

To avoid information loss issues of blind dimensionality reduction (DR) in feature engineering, the entities and relations are clustered prior to training. This is performed to sum up the triple frequency per each cluster for deriving entity or relation feature vectors. This way, the number of features in entity/relation input vectors is reduced to their corresponding number of clusters. In the evaluation section, the clustering method as well as the optimal number of clusters are elaborated. Algorithm 1 demonstrates the clustering approach to feature engineering.

| **Algorithm 1.** Pseudo-code for clustering approach to feature engineering |
|---|

**Input:**
- (head,relation,tail) triples

**Output:**
- feature vector for each entity and relation

**Procedure:**
1. V_max ← 0
2. Assign feature vector of entities to the sparse vector only nonzero in related
   entities and relations[2]
3. For each clustering method in { 'spectral clustering with K-means', 'K-means',
   'Fuzzy C-means', 'K-nearestneighbor'}:
4.     For each kernel in [rbf, sigmoid, polynomial, linear, cosine]:
5.         For each K in {50, 100, 200, 400}:
6.             Cluster feature vectors of entities
7.             V ← Compute variance of all cluster centers (averages)
8.             Replace best clustering setting and V_max , if V> V_max
9. Cluster feature vectors of entities with the best found clustering setting
10. Re-compute feature vectors of entities having length of clusters_count, and
    having nonzero values in indices that entity exist, with intensity of number
    of entities and relations per the cluster
11. V_max ← 0
12. Assign feature vector of relations to the sparse vector only nonzero in related
    entities
13. For each clustering method in { 'spectral clustering with K-means',

---

[2] The entity feature vector is catenation of entity-relation contingencies with entity-entity contingencies. The values of vector for non-related entities and relations are zero. Non-zero values are counted from the number of entities and relations holding the considered entity.

```
       'K-means',
           'Fuzzy C-means', 'K-nearest neighbor'}:
14.      For each kernel in [RBF, sigmoid, polynomial, linear, cosine]:
15.           For each K in {50, 75, 150, 300}
16.                Cluster feature vectors of relations
17.                V ← Compute variance of all cluster
centers (averages)
18.                Replace best clustering setting and
V_max, if V> V_max
19.   Cluster feature vectors of relations with the best found clustering
setting
20.   Re-compute feature vectors of relations having length of clusters
      count, and
        having nonzero values in indices that entity exist, with intensity of
      number
        of entities per the cluster
21.   Return recomputed feature vectors
```

As shown by the proposed fine-tuner in Algorithm 1, we prepared a mechanism to find the best clustering algorithm, number of clusters and the kernel method that transforms the data. The feature vectors (triple frequencies) are passed to a kernel method provided by Scikit-Learn library "pairwise_kernels". This method respectively uses the functions Radial Basis Function (RBF) $k(x_i, x_j) = \exp(-\|x_i - x_j\|^2)$, Sigmoid $k(x_i, x_j) = 1/(1+\exp(-x^T_j x_i))$, Polynomial $k(x_i, x_j) = (x^T_j x_i + 1)$, Linear $k(x_i, x_j) = x^T_j x_i + 1$, and Cosine $k(x_i, x_j) = x^T_j x_i / \|x_j\| \|x_i\|$. In these functions, $x_i$ is a column feature vector corresponding to triple frequency of one entity or relation, and $\|.\|$ is the Euclidean norm. After kernelizing feature vectors, they are passed to the clustering method of interest provided by the fine-tuner algorithm. The best clustering method should find the cluster means that are farthest from each other. We used the variance metric over the sought cluster centers to compare the effectiveness of the different clustering models (clusters discrepancy).

### 3.4 Ranking and Sampling Approach

There are two possible formulations for the target loss function of KGC in order to rank and extract positive tails from negative ones depending on the head and relation input [22]. These two methods are named as point-wise method and list-wise method. Before calculating point average, the point-wise method passes each point score to Sigmoid, while the list-wise method passes all points together to a Softmax function. Softmax pushes values to the extreme and achieves more than Sigmoid by its exponentially scaled reweighting function.

The loss function imposed on rank scores $h_i(e, r)$ for point-wise methods is shown in Eq. 10.

$$L(e, r, y) = -\sum_{i \in \{i|y_i = 1\}} \log(h(e, r)_i) - \sum_m E_{j \sim P_y} \log(1 - h(e, r)_j) \qquad (10)$$

The first term penalizes low positive scores, while the second one penalizes high negative scores. e and r are the input embedding vectors of a training instance in S. We define $y \in R^s$ as a binary label vector for the cases where $y_i = 1$ denotes candidate i as a positive label. The index m counts the negative samples drawn from a sampling distribution. In Eq. 8, $h_i(e, r)$ is sigmoid($W^c_{[1,:]} t + b_s$), where t is defined in Eq. 7.

For the predicted scores to be collected, list-wise method is commonly used in different case studies [12, 23, 24]. The formula for computing list-wise loss function is a cross-entropy over Softmax values, which is described in Eq. 11.

$$L(\boldsymbol{e}, \boldsymbol{r}, \boldsymbol{y}) = -\sum_i^{|y|} \frac{\mathbf{1}(y_i=1)}{\sum_i \mathbf{1}(y_i=1)} \log(h(\boldsymbol{e}, \boldsymbol{r})_i), \qquad (11)$$

where, function $\mathbf{1}(y_i = 1)$ returns 1 when $y_i$ is 1 and 0 otherwise. The positive candidate contains positive probabilities, in contrast to the negative candidates whose probabilities are zero. The process in Fig.1 represents h as a Sigmoid function, following the point-wise ranking scheme. It is flexible and can be changed to Softmax to comply with the list-wise method. The formulation of the above explanation is given in Eq. 12.

$$h(\boldsymbol{e}, \boldsymbol{r})_i = \frac{\exp(W^c_{[i,:]} \tanh(\boldsymbol{e} \oplus \boldsymbol{r}) + \boldsymbol{b_p})}{\sum_j \exp(W^c_{[j,:]} \tanh(\boldsymbol{e} \oplus \boldsymbol{r}) + \boldsymbol{b_p})} \qquad (12)$$

The proposed ProjB method uses margin-based pair-wise ranking loss, which helps the parameters to change simultaneously. This enables the model to feed on the positive samples score, while diminishing the scores of the negative samples. We used a factor to reduce the number of negative samples which improves the effect of positive samples and reduces the time of training [25-27]. Negative sampling has previously been used in plenty of settings such as Word2Vec, Object Detection and several other KGE methods [26, 28, 12].

The following sampling methods are used to address the problem of large-sample-size. These sampling approaches are compared with respect to each other to report accuracy in Table 3.

**Candidate Sampling.** This approach is applied in the ProjE method. It defines a negative candidate sampling rate value to limit the number of negative samples used along with positive ones in the model training stage. The lower the candidate sampling probability is, the lower the extent to which negative triples loss is added to the positive loss. This approach reduces high computational burden ($O(N^3)$) due to regarding all negative triples per each update. The same procedure is followed in ProjE training phase where the candidate sampling rates are tuned and selected from the set $p_y \in \{5\%, 25\%, 50\%, 75\%, 95\%\}$.

**Adaptive Sampling.** This sampling approach keeps track of the prediction accuracy during training stage when a new sample is pushed to the trainer. This approach is designed hoping to speed up the algorithm by training on low-score data more frequently. The first outcome is to prevent embedding vectors from getting caught into local optima, by adaptively increasing randomization. It also feeds lowest scores triples more frequently to the trainer, which raises accuracy in entities/relations that are grouped together. The second outcome that adaptive sampling addresses large-sample-size problem (of graph) by helping to seek for important triples to be fed into the training algorithm.

Table 1. Comparison of ProjB Parameters Count with Other Bilinear and Linear KGEs

| Model | Parameters | Prerequisites |
|---|---|---|
| Unstructured | $n_e k$ | - |
| TransE | $n_e k + n_r k$ | - |
| HolE | $n_e k + n_r k$ | - |
| PTransE | $n_e k + n_r k$ | PCRA |
| ProjE | $n_e k + n_r k + 5k$ | - |
| Proposed ProjB | $k * (n_e + n_r + C_E + C_R + 1)$ | - |
| RTransE | $n_e k + n_r k + q$ | TransE, PCRW |
| LFM | $n_e k + n_r k + 10k^2$ | - |
| SME (bilinear) | $n_e k + n_r k + 2k^3$ | - |
| TransH | $n_e k + 2n_r k$ | - |
| RESCAL | $n_e k + n_r k^2$ | - |
| SE | $n_e k + 2n_r k^2$ | - |
| TransR | $n_e k + n_r (k + k^2)$ | - |
| NTN | $n_e k + n_r (zk^2 + 2zk + 2z)$ | - |

**Weighted Sampling**. A knowledge graph contains complex and nested relations, which are not only provided in a hierarchical structure, but are also identified based on their frequency in connection to all of the heads and tails interactions. Similar to using Term Frequency Inverse Document Frequency (TF-IDF) for detecting informative rare terms, uncommon relations may give more certainty on linking the possible entities [46]. Hence, the triples with rare relations are assigned larger weights to be sampled during the next iterations. This weighting function has been used along with the negative candidate sampling rate, which is formulated as shown in Eq. 13.

$$P_y(H, R, T) = \frac{C \times Level(r)}{\#(h,r,t)_{|r=R} \times \#(Unique(r|h=H)) \times \#(Unique(r|t=T))} \quad (13)$$

In this Equation, where the notations H, R, T represent head, relation and tail respectively for which the sampling probability is computed. C is the normalization constant to make it a probability distribution. The number of triples containing the relation R is shown by $\#(h, r, t)_{|r=R}$ i, and $\#(Unique(r|h = H))$ is the number of unique[3] relations with head H. $\#(Unique(r|t = T))$ is number of unique relations with tail T. $Level(r)$ is the level of relation in the dataset FB15K. As WN18 does not provide level information for relations, this term is set to 1 for WN18.

---
[3] Without repetition

The notation of $\#(h, r, t)_{|r=R}$ is used to penalize sampling triples with highly frequent relations as they might be common and ordinary information with little information content. The term $\#(Unique(r|h = H))$ is used to take into account the diversity of relations with head H in the whole graph. The higher the diversity of relations, the more informative the considered relation is. Terms $\#(Unique(r|h = H))$ and $\#(Unique(r|t = T))$ are the same in homogenous graph and these formulations only apply to heterogeneous graphs. Each ingredient in finding $P_y$ will be calculated, replicated alongside all triples and get produced by each other in a pair-wise way. The complexity order for this computation is $O(N^2)$. In the evaluation analysis section, we show that this sampling approach achieves the highest accuracy over the benchmark KGs compared to the other sampling approaches.

### 3.5 Complexity Analysis

In order to emphasize on the advantages of ProjB, we provide a parameter comparison in Table 1. Here we carefully analyzed a number of other models in terms of their parameters. This list cover most of the state-of-the-art KGE model that use translation in their score function, or semantic matching methods. We also covered other types of models such as Unstructured, LFM, and NTN. The third column is about the prerequisites of the models that is only valid for PTransE and RTransE.

The comparison shows that the number of parameters for ProjB is not causing any complexity in computations and instead affects the speed positively. According to Table 1, the number of parameters for ProjB merely equals to that of ProjE. Despite having smaller number of parameters than all the bilinear methods and most linear ones, the KGC accuracy of ProjB exceeds all other approaches.

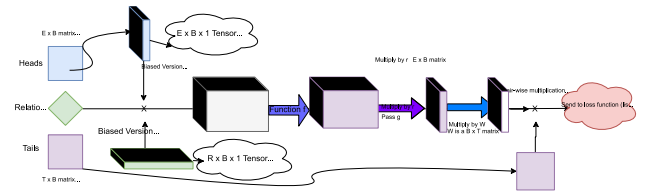

*Figure 2: Proposed Schematic of Algorithm 2, for Parallel-processing Objectives*

In the testing and validation phase, only tuples of head-entity indices are passed to the model. The model extracts the engineered features out of the associated head and relation. This is done in order to find model gradients and update embedding vectors.

Table 2. statistical analysis for testing the relative number of local optimas in ProjB versus ProjE

| Over 50 trials | FBK15 (Raw) | FBK15 (filtered) | WN18 (Raw) | WN18 (filtered) | Upper bound |
|---|---|---|---|---|---|
| Mean (over 50 trials) of average loss divisions ( CE(ProjB)/ CE(ProjE) ) | 0.6854 | 0.7882 | 0.8466 | 0.8503 | 0.8797 |
| Std (over 50 trials) of average loss divisions | 0.83 | 0.73 | 1.62 | 0.29 | 0.98 |
| P-value | 0.197 | 0.226 | 0.049 | 0.363 | 0.146 |
| Trustworthy (Meaningful) | **Yes** | **Yes** | **No** | **Yes** | **Yes** |

## 3.6 Statistical Analysis of Tail Embedding Vectors in ProjE Caught in Local Optima

**Algorithm 2.** Pseudo-code for Batch-processing Mode of ProjB

**Input:**
- Batch count
- Triple indices in the batch
- Entity-relation co-occurrence matrix

**Output:**
- Loss function pipeline to be optimized

**Procedure:**
1: For each triple, seek for three vectors in Entity-relation co-occurrence matrix; i.e., row vectors associated with desired head and tail index, and column vector of the concerning relation index; add each vector column-wise to heads, tails, and relations respectively.
2: Add a third dimension to heads and relations matrices to make them 3D tensors. First and second dimensions of tensors are assumed to be the first and second dimensions of heads and relations row vectors.
3: Do a pairwise multiplication of heads tensor by $D_e$, and tails by $D_r$ from Eq. 6.
4: Add to the weighted tensors (heads and relations) element, learnable biases $b_{p_{c_e}}$ and $b_{q_{c_r}}$.
5: Using tensor multiplication API, multiply resulted heads tensor from the 2X3 side by the resulted relations tensor from the 1X3 side; leaving a 3D tensor of heads-batches-relations.
6: Pass the resulted tensor to f (sigmoid) function.
7: Multiply 2X3 side of the resulted tensor by 1X2 side of relations tensor to generate batch vectors equivalent to formula t in Eq. 7.
8: Pass the resulted matrix as '$t$' to $g(W_{i,:}t + b_p)$.
9: Compute loss function value either using Eq. 8 or Eq. 9.

To show that embedding vectors of tails derived by ProjE may end up in local optima, we compare cross entropy loss of ProjE with our proposed ProjB in 50 sets of randomized triples. In 50 different trials, we randomly sample triples from KG with a uniform distribution up to the number of all possible triples. As sampling is performed with replacement, we expect that on average 66 percent of the observations are present in each trial.[4] We compare the cross-entropy loss value of ProjE with the one in ProjB. ProjB loss is divided by ProjE loss. The null hypothesis is that such division is above 1 generally. Table 2 shows that the hypothesis is rejected for three cases out of the four database cases. Only in one case the P-value got under 0.05 and the null hypothesis was accepted. The trustworthy values of 'Yes' manifest rejection of null hypothesis. Therefore, ProjE generally gets caught in local optima more often than our proposed ProjB. Table 2 tests meaningfulness of the claim that ProjE relatively gets caught in local optima (∼ cross entropy loss of ProjB is generally lower than ProjE). CE is acronym for Cross-Entropy. CE(.)= CE($h_i$(e, r), $ĥ_i$(e, r)) where h is estimated score for tail in triple, and $ĥ$ is target score per each tail.

## 3.7 ProjB Optimization Prunes Unnecessary Dimensions and Adds New Interactions

As discussed in Section 2, $t = f(D_e e + D_r r + b_c)$ in ProjE is replaced with Eq. 7. e and r are the column-wise vectors. As each dimension of e is added to only one specific dimension of r, wrong update of one dimension of e leads to wrong update of the corresponding dimension of r too. Chance of wrong update of one entity dimension in our proposed formulation as given in Eq. 7 gets reduced because of more interactions among greater number of dimensions. To show that, we expand Eq. 7 to Eq. 14.

$$t = f\left(\left(D_e e + b_{p_{c_e}}\right)\left(D_r r + b_{q_{c_r}}\right)^T\right) r = f\left(D_e e r^T D_r + b_{p_{c_e}} D_r r^t + D_e e b_{q_{c_r}}^T + b_{p_{c_e}} b_{q_{c_r}}^T\right) r \qquad (14)$$

The first term in f input shows interaction of every single dimension of e with all dimensions of $r^T$. This means that, wrong update of one dimension has the chance to be corrected in numerous gradient updates of every single dimension of r. As another r is added to the right side of t in Eq. 7, the optimizer has the opportunity to prune unnecessary dimension of relation by fine-tuning the embedding vectors in specific way. By piecewise-linearizing f over operating point, we see the interactions as shown in Eq. 15 after expanding t.

---
[4] The reason for not selecting all triples is that the graph is incomplete and some triples are missing from it.

$$t \sim D_e e r^T D_r r + b_{p_{c_e}} D_r r^t r + D_e e b_{q_{c_r}}^T r + b_{p_{c_e}} b_{q_{c_r}}^T r + C \quad (15)$$

We realize that compared to Eq. 5, two more interactions are added to our formulation: 1) entity-relation interactions (first term), and 2) relation-relation interactions. Therefore, our model is robustly capable of handling greater number of interactions between entities and relations.

## 4 Evaluation

We evaluate our proposed ProjB model on the entity prediction task over known benchmarks namely FB15K and WN18 knowledge graphs. The purpose of the evolutions is to show the outperforming performance of the model in comparison to other linear, bilinear models as well as the base ProjE method. We will discuss the information regarding implementation, the prediction task, underlying knowledge graphs, and the parameter tuning. Then the effects of clustering and sampling the entity and relation are discussed. The last part of this section is devoted to a discussion regarding the scalability and speed of ProjB caused by Tensor-based pseudo-code that was discussed in Alg. 2. In addition, we further show the effect of the loss function in optimizing the combination of cross entropy between target $\hat{h}_i(e, r)$ vs $h_i(e, r)$, and regularization of the embedding cluster. We shall note that as the objective of this work is merely to improve translative and bilinear approaches, we considered the discussion about the other neural-based state-of-the-art algorithms out of scope.

### 4.1 Evaluation Setup

The evolutions of this paper have been performed over two known benchmarks namely FB15k and WN18. The FB15k dataset is a subset of FreeBase knowledge graph that contains 14951 entities, 1345 relationships, 483142 triples in train, 50,000 triples in validation, and 59071 triples in the test dataset. WN18 is a version of the WordNet dataset that has 117,000 subsets with %70 of it as training data. One of the evaluation metrics is Hits@10 for entity prediction that is the number of tails given a head and relation, aggregated over all head-relation tuples. The percentage of correct tails out of top 10 high-scored detections is the required score for this task. In terms of model selection, the number of iterations have been set to 100, while the batch size is selected as 30, empirically fine-tuned from the set {10, 30, 60, 100} based on the best result. The embedding dimension, which is also equal to the number of clusters, is set to 100 for entities and 75 for relations. These parameters are fine-tuned out of multiple executions on the dimension set of {50, 100, 200, 400} for entities and {50, 75, 150, 300} for relations. For the clustering method, K-means is selected for grouping entities/relations. This is needed for computing entity co-occurrence features. The best accuracy was achieved using K-means, among multiple clustering techniques that are tried such as K-means, Spectral Clustering, Fuzzy C-means, and K-nearest Neighbors. We implemented multiple kernel types such as Nearest-Neighbor (NN), Polynomial, and Spectral-Clustering. NN kernel in K-means provided higher validation accuracy. Adam optimizer [29] is used for entity prediction task where tuned hyper-parameter settings are $\beta_1 = 0.8$, $\beta_2 = 0.99$, and $\varepsilon = 1e-8$. The reason for that choice is its outperformance with respect to AdaGrad, RMSProps, Momentum and simple Gradient method. For the purpose of fair evaluations and comparison set up, the same hyper-parameter tuning is chosen for each one. Weight decay was added to the underlying optimizer in order to account for fluctuations in relation switching. This is according to the best practices that it generally improves the KGC prediction accuracy [13].

### 4.1 Results and Discussion

Table 2 demonstrates the results of the three types of KGE models, namely 1) translation-based, 2) bilinear, and 3) network-based methods with their entity prediction accuracy over the chosen KGs namely FB15K and WN18.

Moattari et al.The translation-based methods account for the variants of TransE, all of which having transformative relation between entities. The set of methods considered in the bilinearity type has multiplicative interactions between embedding vectors in the model. The last type are mostly the generic models that contain combinatorial or neural structures. As it is evident in Table 3, the detection rate of ProjB outperforms all the bilinear methods. The use of biases $b_{p_{c_e}}$ and $b_{q_{c_r}}$ have determining roles in reinforcing bilinear mechanism for part of database over linear transformation.

Table 3. Comparison of ProjB vs. Existing Translation-based, Bilinear, and NN-based models.

| Type | Models | Energy Function | Raw | Filtered | Raw | Filtered |
|---|---|---|---|---|---|---|
| Translation | Unstructured (Bordes et al., 2013) | [1] | 4.5 | 6.3 | 35.3 | 38.2 |
| Translation | PTransE-MUL (Lin et al., 2015a) | TransE with relation paths [36] | – | 77.7 | – | – |
| Translation | PTransE-RNN (Lin et al., 2015a) | TransE with relation paths [36] | 50.6 | 82.2 | – | – |
| Translation | PTransE-ADD (Lin et al., 2015a) | TransE with relation paths [36] | 51.4 | 84.6 | – | – |
| Translation | TransSparse (Ji et al., 2016) | $\left\|\left\|M_r^h(\theta_r^h)h + r - M_r^t(\theta_r^t)t\right\|\right\|_{l_{1/2}}^2$ | 53.5 | 79.5 | 80.1 | 93.2 |
| Translation | TransH (Wang et al., 2014) | $\left\|\left\|(h - \omega_r^T h \omega_r) + d_r - (t - \omega_r^T t \omega_r)\right\|\right\|_2^2$ | 45.7 | 64.4 | 73 | 86.7 |
| Translation | TransE (Bordes et al., 2013) | $\left\|\left\|h + r - t\right\|\right\|_{l2}$ | 34.9 | 47.1 | 75.4 | 89.2 |
| Translation | STransE (Nguyen et al., 2016b) | $\left\|\left\|W_{r,1}h + r - W_{r,2}t\right\|\right\|_{l_{1/2}}$ [37] | 51.6 | 79.7 | 80.9 | 93.4 |
| Translation | RTransE (Garcia-Duran et al., 2015) | Recurrent form of TransE [38] | – | 76.2 | – | – |
| Translation | TransD (Ji et al., 2015) | $\left\|\left\|M_{rh}h + r - M_{rt}t\right\|\right\|_2^2$ | 53.4 | 77.3 | 79.6 | 92.2 |
| Translation | TransR (Lin et al., 2015b) | $\left\|\left\|hM_r + r_c - tM_r\right\|\right\|_2^2$ | 48.2 | 68.7 | 79.8 | 92 |
| Translation | CTransR (Lin et al., 2015b) | $\left\|\left\|hM_r + r_c - tM_r\right\|\right\|_2^2 + \propto \left\|\left\|r_c - r\right\|\right\|_2^2$ | – | 70.2 | – | 92.3 |
| Bilinearity | SME bilinear (Bordes et al., 2012) | $(W_{u1}r + W_{u2}h + b_u)^T(W_{v1}r + W_{v2}t + b_v)$ | 31.3 | 41.3 | 65.1 | 74.1 |
| Bilinearity | ANALOGY (Liu et al., 2017) [30] | Min E $[v_s^T w_r v_o \| y]$ | – | 85.4 | – | 94.7 |
| Bilinearity | DistMult (Yang et al., 2014) | Eq. 3 | – | 82.4 | – | 93.6 |
| Bilinearity | DistMult (Yang et al., 2015) | Eq. 3 | – | 89.3 | – | 94.6 |
| Bilinearity | NTN (Socher et al., 2013) | $u_r^T \tanh(h^T W_r t + W_{rh}h + W_{rt}t + b_r)$ | 53 | 41.4 | – | 66.1 |
| Bilinearity | RESCAL (Nickel et al., 2011) | [34] | 28.4 | 44.1 | – | 52.8 |
| Generic | TKRL (Xie et al., 2016) | $\left\|\left\|M_{rh}h + r - M_{rt}t\right\|\right\|_2^2$ | 50.3 | 73.4 | – | – |
| Generic | SE (Bordes et al., 2011) | $\left\|\left\|R_u h - R_u t\right\|\right\|_{t1}$ | 28.8 | 39.8 | 68.5 | 80.5 |
| Generic | LFM (Jenatton et al., 2012) | [35] | 26 | 33.1 | – | 81.6 |
| Generic | SimplE (Kazemi and Poole, 2018) | $\frac{1}{2}(v_{h,1}^T W_r v_{t,2} + v_{t,1}^T W_{r^{-1}} v_{h,2})$ [39] | – | 83.8 | – | 94.7 |
| Generic | TorusE (Ebisu and Ichise, 2018) | [40] | – | 83.2 | – | 95.4 |
| Generic | ER-MLP (Dong et al., 2014) | [41] | – | 80.1 | – | 94.2 |
| Generic | ConvE (Dettmers et al., 2018) [42] | $v_t^T g(vec(g(concat(v_h, v_r) * \Omega))W)$ | – | 87.3 | – | 95.5 |
| Generic | RotatE (Sun et al., 2019) [43] | $-\left\|\left\|c_h \circ c_r - c_t\right\|\right\|_{l_{1/2}}; c_h, c_r, c_t \in C^k$ | – | 88.4 | – | 95.9 |
| Generic | HolE (Nickel et al., 2016) | $r^T(h * t), *:$ circular correlation | – | 73.9 | – | 94.9 |
| Baseline | ProjE_pointwise (Shi et al., 2017) | Eq. 5, Eq. 10 [12] | 56.5 | 86.6 | – | – |
| Baseline | ProjE_listwise (Shi et al., 2017) | Eq. 5, Eq. 11 [12] | 54.6 | 71.2 | – | – |
| Baseline | ProjE_wlistwise (Shi et al., 2017) | Eq. 5, Eq. 11 [12] | 54.7 | 88.4 | – | – |
| Proposed | ProjB_pointwise | Eq. 6, Eq. 7, Eq. 10 | 61.9 | 88.9 | 82.1 | 96.4 |
| Proposed | ProjB_listwise | Eq. 6, Eq. 7, Eq. 11 | 62.2 | 90.3 | 82.5 | 97.2 |

The results, caused by the proposed cluster dependent approach, encodes more localized information suitable to infer the right tails. ProjB also outperforms over ProjE by at least 7% over FB15k that shows the initial research hypothesis is achieved as our model resolves the problems of ProjE (see Section 1). In addition, ProjB succeeded in outperforming existing translation-based models, which approves that linearity and mere translations cannot handle nonlinearities between entities and relations. Therefore, our learnable attention mechanism over all entity-relation dimensional interactions has helped to learn the nonlinearity inherent in the KG's structure. The results are over test data which are unseen by hyper-parameter tuner.

### 4.1.1 Behavior of Center Variance in Clusters During Training

Fig. 3 shows the behavior of clusters center variance for entity and relation embedding vectors over FB15K. The left side of the figure for parts (a) manifests the variance for relation centers of clusters, while the right side for part (b) represents the results for entities. Results show that in all the cases, embedding vectors get dissipated, having the clusters taken distance from each other. The clusters of relations more smoothly deviate from each other compared to corresponding entities. Furthermore, the final relative variance for relations is higher than its counterpart for entities. A deduction out of these observations can be that entities are more inter-related, hence making their clusters relatively closer to each other compared to relations. Another important fact to point out is the characteristic of the underlying KG. For example, the WN18 KG has more abrupt cluster deviation growth rate than that of FB15K. It seems that, the more comprehensive a KG is, the more interrelated its entities and relations are. This generally leads to partially closer clusters.

## 4.2 Comparisons of Feature Engineering, Sampling and Pre-processing

We provide insights as to how our data processing approaches may have contributed to the performance improvement of the proposed ProjB model. We applied simple feature engineering and sampling methods. The feature engineering methods described here can be used as easy to use frequency data instead of pre-trained data that demands a huge number of computations. Table 4 shows comparative performance of our proposed ProjB model for different feature engineering, data sampling, and pre-processing use case scenarios. From our experimental study, we only show Knowledge Graph Completion detection rates for list-wise approach as it gave the best performance as shown in Table 3. The reason to substantiate each comparison has been elaborated upon hereunder.

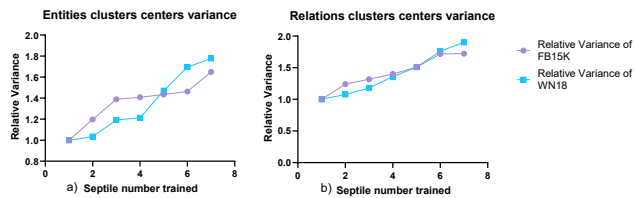

**Figure 3: Cluster Center Variance. The x axis represents the data septile being trained and the y axis depicts the relative variance growth of all clusters during training.**

### 4.23 Effects of Speed Improvement by Batch Processor

To assess the effect of using tensor processing API, Alg. 2 has been evaluated and compared in the task of KGC over FB15k knowledge graph. For the batch sizes greater than 1, Alg. 2 is used and the training time is recorded for each. The system used for training has a Tesla4 GPU with 15GBytes of RAM under Linux OS. For the cases without batches (shown as batch size of 1 in Fig. 4), the training time was the highest of all with a salient difference. It verifies the power of the proposed algorithm to reduce the training duration to such a high extent. As Fig. 4 shows, the elapsed time decreases with an increasing slope. However, due to the fact that beyond batch size of 30, the accuracy undergoes a decay, the batch size was set to 30. Overall, the scalability of the algorithm will still be maintained, as the decay in accuracy is not substantial.

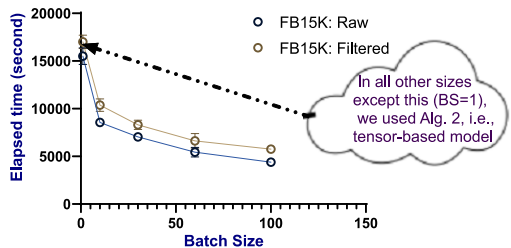

**Figure 4: Training elapsed time for KGC task based on the number of batches**

## 4 Conclusion and Future Work

A new variation of ProjE graph embedding model, named ProjB, is proposed in this paper. The model not only applies to bilinear interactions, but also propose more holistic bilinear model. Other available contributions include the model's ability to learn nonlinearity as required by KGEs using an attention mechanism, and consider both translational and bilinear aspects by adding relation and entity cluster-specific embedding vectors. Results from experimental evaluations

Moattari et al.

Table 4. Test entity prediction Hit@10 detection rate for comparing influential settings in KGE problems. BS accounts for batch size. DR means dimensionality reduction. FE is feature engineering.

| ProjB: Listwise | Specification | Sec. 3.2: Effect of FE Clustering | | Sec. 4.2.1: Cluster Update Effect in Training | | Sampling approach effects | | |
|---|---|---|---|---|---|---|---|---|
| | Training specification | No use: PCA DR | Use: Cluster | No Updates | Adaptive update | Negative Candidate Sampler | Weighted Sampler | Adaptive Sampler |
| **FB15K** | BS=1 | **57.3** | 56.3 | 58.3 | **58.6** | 56.3 | **57.0** | 56.3 |
| **Hits@10** | BS=10 | 55.3 | **57.4** | 54.4 | **55.1** | 53.2 | **56.2** | 55.4 |
| **Raw** | BS=30 | 55.9 | **62.2** | 54.6 | **62.2** | 55.1 | **62.2** | 61.6 |
| **FB15K** | BS=1 | 76.6 | **82.8** | 75.6 | **80.9** | 75.5 | **83.7** | 81.9 |
| **Hits@10** | BS=10 | 85.1 | **88.1** | 87.4 | **89.7** | 87.5 | **89.9** | 88.3 |
| **Filtered** | BS=30 | 86.2 | **90.3** | 86.6 | **90.3** | 88.4 | 90.3 | **90.8** |

show that our model outperforms the state-of-the-art translational and bilinear models in Knowledge Graph Completion task. In addition to the main KGE model, a tensor based model for speed improvement is formulated and implemented in the proposed approach and the speed results are provided. Finally, the new sampling and simplistic feature reduction approaches have been introduced as helpful tools for the researchers. The complexity and number of problems are relatively close to the baseline ProjE model.

In this work, we focused on improving over linear and bilinear models; adapting the ProjB beyond these is a future work. Therefore, the current version of ProjB is not comparable to models such as IRN and ComplEx-N3 [31, 32] due to their high number of required parameters and being out-of-scope[5] for this study. It seems that state space based models (e.g., memory oriented approach in IRN) and nonparametric data driven models (e.g., tensor decomposition in ComplEx-N3) are vital for the improvement of KGC tasks. In the prospective works, new regularizations will be injected to ProjB to capture nests of entities interrelationships using state space constraints. Another approach is to formulate a higher order tensor decomposition loss to learn deep data-driven dictionaries responsible for three-ways interaction of relations. These are planned as the next steps to improve ProjB model as a general framework.

---

[5] Scope of this work is to merely improve over linear and bilinear models. Neither IRN nor ComplEx-N3 have these characteristics. However, they are important in the sense of dictionary atoms that can easily improve distributed models using rules, clusters and conditions. Especially when a lot of improvements in the proposed work are due to adaptive and initialized clusters and rules.